\documentclass[sigconf]{acmart}

\usepackage{booktabs} 

\setcopyright{rightsretained}
\usepackage{amsmath}
\usepackage{mathrsfs}
\usepackage{mathtools}
\usepackage{relsize}
\usepackage{graphicx}
\usepackage{float}
\usepackage{paralist}
\usepackage[T1]{fontenc}
\usepackage[lined,boxed,commentsnumbered]{algorithm2e}

\begin{document}
\title{A Short Survey of Biomedical Relation Extraction Techniques}

\author{Elham Shahab}
\affiliation{%
  \institution{Department of Computer Engineering\\ Islamic Azad University, Yazd Branch}
}
\email{ma.shahab@iauyazd.ac.ir}

\begin{abstract}
Biomedical information is growing rapidly in the recent years and retrieving useful data through information extraction system is getting more attention. In the current research, we focus on different aspects of relation extraction techniques in biomedical domain and briefly describe the state-of-the-art for relation extraction between a variety of biological elements.
\end{abstract}

%
%

\keywords{Information Extraction, Biomedical text mining, Relation Extraction.}


\maketitle

\section{Introduction}
Biomedical literature is growing rapidly, Cohen and Hunter in \cite{cohen2008getting}  explain how the growth in PubMed/MEDLINE publications is phenomenal, which makes it  a potential area of research with respect to information and data mining techniques.In fact, it is quite difficult for biomedical scientists to adjust new publications and come up with relevant publications in their own research area. To address this, text mining and knowledge discovery is getting more attention these days in biomedical sciences. In fact, automated text processing methods try to overcome text information overload and transform those data into machine understandable format. Text mining and knowledge extraction techniques along with statistical machine learning algorithms are widely used in medical and biomedical domain such as \cite{pouriyeh2017comprehensive, yoo2012data}. In particular, text mining methods have been applied in a variety of  biomedical branches and domains such as gene clustering, protein structure prediction, clinical diagnosis, biomedical hypothesis and etc. In this section, we briefly describe some of the relevant research in biomedical domain and explain some of the state-of-the-art relation extraction techniques with respect to data mining approaches in biomedical discipline.

\section{Relation Extraction}
Determining the relationships among biomedical entities is the key point in relation extraction in Biomedical domain. The ultimate goal is to locate the occurrence of a specific relationship type between given two entities. There are lots of extraction format available in biomedical domain such as RDF \cite{goble2008state, newman2008scale} and XML format \cite{doroodchi2009investigation,bales2005framework, pouriyeh2010secure} which is widely used. For instance, in the genomic area, extracting interactions between genes and proteins such as gene-diseases or  protein-protein relationships is very important and getting more attention these days. Relation extraction is usually integrated with the similar challenges as NER, such as creation of high quality annotated data for training and assessing the performance of relation extraction systems.
There are different text mining techniques \cite{allahyari2017brief} such as topic modeling \cite{allahyari2016tagging,allahyari2015automatic}, information extraction \cite{wimalasuriya2010ontology,2017arXiv170607992T}, text summarization \cite{allahyari2017text},  and clustering \cite{fodeh2011ontology, allahyari2017brief} for relation extraction between some of the different types of biological elements such as genes, proteins and diseases that will be discussed in the following sections.

\section{Biomedical Relation Extraction Techniques}
Knowledge and Information extraction and in particular relation extraction tasks have widely studied various biomedical relations. There are lots of ongoing research in biomedical relation extraction due to critical roles of genes and proteins interactions in different biological processes. Many different approaches for biomedical relation extraction have been proposed which can be a simple systems that only rely on co-occurrence statistics to complex ones which use syntactic analysis and dependency parse trees. The entities co-occur based technique is considered as a the most straightforward technique which is based on this fact that If they mentioned together more frequently, there is a chance that they might be related together in some way. For example, Chen et al. \cite{chen2008automated} introduce a co-occurrence statistics method to calculate and evaluate the degree of association between disease and relevant drugs from clinical narratives and biomedical literature. 
An other approach in this area is Rule-based approaches. In this technique a set of methods used for biomedical relation extraction. Usually, rules are defined manually by domain experts \cite{vsaric2006extraction} or automatically generated by using machine learning methods \cite{hakenberg2005lll} from an annotated corpus. Hakenberg et al. \cite{hakenberg2005lll} define and extract syntactical patterns learned from labeled examples and match them against arbitrary text to detect protein-protein interactions.  Classification-based techniques are also widely used methods for relation extractions in biomedical domain \cite{2017arXiv170607992T}. For example, Rink et al. \cite{rink2011automatic} identify a set of features from multiple knowledge sources such WordNet and Wikipedia. In the next phase train and then apply a supervised machine learning technique, Support Vector Machine (SVM), to extract the relations between medical records and treatments. In addition, Bundschus et al. \cite{bundschus2008extraction} have applied a supervised machine learning method that detects and classifies relations between diseases and treatments extracted from PubMed abstracts and between genes and diseases in human GeneRIF database. 

Relation extraction methods improved fundamentally by considering the syntactic and semantic structures. Specifically, syntactic parsing methods, including dependency trees (or graphs) are able to produce syntactic information about the biomedical text which reveals grammatical relations between words or phrases. For example, Miyao et al. \cite{miyao2009evaluating} conducted a comparative of several state of the art syntactic parsing methods, including dependency parsing, phrase structure parsing and deep parsing to extract protein-protein interactions (PPI) from MEDLINE abstracts.

Having faced the increasing growth of biomedical data, many approaches utilized machine learning techniques to extract useful information from syntactic structures rather than applying manually derived patterns \cite{simpson2012biomedical}. Airola et al. \cite{airola2008graph} propose an all-path graph kernel to calculate the similarity between dependency graphs, and then use the kernel function to train a support vector machine to detect protein-protein interactions. Miwa et al. \cite{miwa2009protein} describe a method to combine kernels and syntactic parsers for PPI extraction. Furthermore, Kim et al. \cite{kim2008kernel} introduce four genic relation extraction kernels defined on the shortest dependency path between two named entities.

Semantic role labeling (SRL), a natural language processing technique that identifies the semantic roles of these words or phrases in sentences and expresses them as predicate-argument structures, is also useful when it is complemented with syntactic analysis. \cite{thompson2009construction,tsai2007biosmile} are examples which have used SRL.

In the following, We describe some of works done for relation extraction between a variety of biological elements.

\subsection{Gene-Disease}
Chun et al. \cite{chun2006extraction} describe a classification-based approach for relation extraction. First they use a dictionary-based longest matching technique which extracts all the sentences that include at least one pair of gene and disease names. Then, they apply a Maximum Entropy-based NER to filter out false positives produced in previous stage. They reach the precision of $79\%$ and recall of $87\%$ which significantly outperforms previous methods. Bundschus et al. \cite{bundschus2008extraction} also propose a classification-based method, Conditional Random Field (CRF), to identify and classify relations between diseases and treatments and relations between genes and diseases. Their system utilizes supervised machine learning, syntactic and semantic features of context. For more information, see \cite{faro2009discovering,tsai2009hypertengene,rindflesch2003semantic,baud2003improving}.

\subsection{Gene-Protein}
Fundel et al. \cite{fundel2007relex} use Stanford Lexicalized Parser to create dependency parse trees from MEDLINE abstracts and complement this information with gene and protein names obtained from ProMiner NER system \cite{hanisch2005prominer}. Then the system applies a few different relation extraction rules to identify gene-protein and protein-protein interactions. They achieved better precision and F-measure and significantly outperformed previous approaches. Saric et al. \cite{vsaric2006extraction} present a rule-based method to extract gene-protein relations. They integrate NLP techniques to preprocess and recognize named entities (e.g. genes and proteins), then apply a separate grammar module, combining syntactic properties and semantic properties of the relevant verbs, to extract relations. Some other works include \cite{coulet2010using,koike2005automatic}.

\subsection{Protein-Protein}
Raja et al. \cite{raja2013ppinterfinder} introduce a system called PPInterFinder to extract human protein-protein interactions (PPI) from MEDLINE abstracts. PPInterFinder integrates NLP techniques (Tregex for relation keyword matching) and a set of rules to identifies PPI pair candidates and then apply a pattern matching algorithm for PPI relation extraction. \cite{madkour2007bionoculars} presents a statistical unsupervised method, called BioNoculars. BioNoculars uses a graph-based method to construct extraction patterns for extracting protein-protein interactions. 
\cite{tikk2010comprehensive} performs a comprehensive benchmarking of nine different methods for PPI extraction that utilizes convolution kernels and confirms that kernels using dependency trees generally outperform kernels based on syntax trees. Similarly, \cite{zhou2008extracting} describes various methods for PPI extractions. For more approaches, see \cite{bui2011hybrid,kim2010walk,airola2008graph,rosario2005multi}.

\subsection{Protein-Point mutation}
The problem of point mutation extraction is to link the point mutation with its related protein and organisms of origin. Lee et al. \cite{lee2007automatic} introduce Mutation GraB (Graph Bigram), that detects, extracts and verifies point mutation from biomedical literature. They test their method on 589 articles explaining point mutations from the G protein-coupled receptor (GPCR), tyrosine kinase, and ion channel protein
families, and achieve the F-score of $79\%, 72\%$ and $76\%$ for the GPCRs, protein tyrosine kinases and ion channel transporters respectively.

A few other algorithms have been developed for point mutation extraction. Rebholz-Schuhmann et al. \cite{rebholz2004automatic} introduce a method called MEMA that scans MEDLINE abstracts for mutations. Baker and Witte \cite{baker2004enriching,baker2006mutation,witte2005combining} describe a method called Mutation Miner that integrates point mutation extraction into a protein structure visualization application. \cite{horn2004automated} presented MuteXt, a point mutation extraction method applied to G protein-coupled receptor (GPCR) and nuclear hormone receptor literature. \cite{doughty2011toward} describes a automatic method for cancer and other disease-related point mutations from biomedical text.

\subsection{Protein-Binding site}
Ravikumar et al. \cite{ravikumar2012literature} propose a rule-based method for automatic extraction of protein-specific residue from the biomedical literature. They use linguistic patterns for identifying residues in text and then apply a graph-based method (sub-graph matching \cite{liu2010biological}) to learn syntactic patterns corresponding to protein-residue pairs. They achieved a F-score of $84\%$ on an automatically created dataset and $79\%$ on a manually annotated corpus and outperforms previous methods. Chang et al. \cite{chang2006protemot} describe an automatic mechanism to extract structural templates of protein binding sites from the Protein Data Bank (PDB). For more information about binding of other ligands to proteins, see \cite{leis2010silico}.

\subsection{Other Types of Interactions}
Recently, there has been an increasing attention to the more complex task of identifying of nested chain of interactions (i.e \emph{event extractions}) rather than identifying binary relations. Because biomedical events are usually complex, effective event extraction normally needs extensive analysis sentence structure. Deep parsing methods and semantic processing are specially very helpful due to the capability of examining both syntactic as well as semantic structures of the biomedical text. For an overview of the currently available methods, see \cite{ananiadou2010event}.

Event extraction has started to be widely used for annotation of biomedical pathways, Gene Ontology annotation and the enhancement of biomedical databases \cite{simpson2012biomedical}. For example, \cite{friedman2001genies} presents a NLP-based system, GENIES, to extract molecular pathways from biomedical literature. 

There are several corpora in the biomedical domain that have integrated event annotations such as BioInfer corpus \cite{pyysalo2007bioinfer}. GENIA Event Corpus \cite{kim2003genia} and the Gene Regulation Event Corpus \cite{thompson2009construction} are other annotated event corpora which are widely employed in biomedical text mining. For a comprehensive overview of the biomedical event extraction and evaluation, see \cite{simpson2012biomedical,ananiadou2010event}.

In addition, there are studies for identifying drug-drug interaction (DDI) in biomedical text. DDI can occur when two drugs interact with the same gene. Percha et al. \cite{percha2012discovery} use a NLP technique \cite{coulet2010using} to identify and extract gene-drug interactions and propose a machine learning technique to predict DDIs. Some other works for DDI are \cite{plake2011computational,garten2010recent,tari2010discovering}.

\section{Discussion}
Although relation extraction between various biological elements (e.g. genes, proteins and diseases) from biomedical literature has attained extensive attention recently, yet these text mining techniques have not been applied to extract relations between other types of molecules, particularly complex macromolecules to these important biological processes (e.g. glycan-protein interactions). The potential reasons of why extracting carbohydrate-binding proteins relationship from biomedical text have almost remained untouched, are as follows:

\begin{enumerate}
	\item Raman et al. \cite{raman2005glycomics} explains that the progress of glycomics has coped with distinctive challenges for developing analytical and biochemical tools to investigate glycan structure-function relations compared to genomics or proteomics. Glycans are more varied in terms of chemical structure and information density than DNA and proteins. In other words, carbohydrate-binding proteins are greatly heterogeneous in terms of their sequences, structures, binding sites and evolutionary histories \cite{doxey2010structural}. This complicates the development of analytical techniques to \emph{accurately} define the structure of glycans which accordingly makes the investigation and understanding of glycan-protein relations difficult. Therefore, the amount of knowledge in this domain is not comparable to genomic area where, it has led to less concentration on this field.

	\item In comparison with genomic area, the glycan-related knowledge bases (e.g. ontologies, databases, etc) which can be used as \emph{background knowledge} to analyze the biomedical literature for information extraction is very restricted in terms of \emph{quantities} and \emph{qualities}. As we explained before (section 1 and 3), there exist many different ontologies and corpora about genes, diseases and proteins which are widely used in text mining, but there are barely a few ones for glycobiology research. For example, UniCarbKB\footnote{\texttt{http://www.unicarbkb.org/}} is a knowledge base and a framework that includes structural, experimental and functional data about glycomic experiments.  Consortium for Functional Glycomics (CFG)\footnote{\texttt{http://www.functionalglycomics.org/}}, funded by US National Institute of General Medical Sciences, is another collaborative effort which facilitates access to databases and services about glycomic research. However, according to previous reason, the algorithms used to automatically produce glycan structures are laborious and not quite accurate which may result in lower quality information whereas the databases and ontologies in genomic area contain curated data. Also, the amount of knowledge in glycobiology research area is extremely small in terms of number of concepts and relations and instances in ontologies and/or the volume of data in databases as opposed to the fairly rich ontologies about genes, proteins and diseases \cite{campbell2013unicarbkb}. 

\end{enumerate}

\section{Conclusion}
Nonetheless, glycoproteomics is an emerging research area and there are many interesting future directions regarding information extraction and knowledge discovery in this domain. Glycoproteomics literature is barely touched by text mining community (due to aforementioned reasons), thus, there is a great demand for creating curated and high quality ontologies for glycoscience information. As we mentioned, UniCarbKB is an example of such systems. Even though, UniCarbKB provides critical information, it really is a database, not an ontology. Additionally, it does not contain a large amount of information. However, UniCarbKB research group has recently started to represent the data in RDF to unify the content and also begun to extend it to encompass more knowledge \cite{campbell2013unicarbkb}.

Another interesting direction is not only to create ontologies, but also to integrate them to invaluable existing ontologies in genomic area and linked open data which is very beneficial, because:
\begin{inparaenum}[\itshape 1\upshape)]
\item Although different ontologies contain different set of concepts, they have inter-relations to each other. This makes new interesting discoveries of hypotheses as well as relation extractions where it would not be possible using ontologies individually.
\item It facilitates the development of various applications for knowledge discovery (e.g. faceted browsing, data visualization, etc) in this domain.
\end{inparaenum}

  There are other interesting research directions in the area of knowledge discovery from glycoproteomics literature, and our propositions are barely scratching the surface.

\bibliographystyle{ACM-Reference-Format}
\bibliography{compbib} 

\end{document}